\documentclass[conference]{IEEEtran}
\IEEEoverridecommandlockouts

\usepackage{cite}
\usepackage{amsmath,amssymb,amsfonts}
\usepackage{bm}  
\usepackage{algorithmic}
\usepackage{graphicx}
\usepackage{textcomp}
\usepackage{xcolor}
\usepackage{siunitx}
\usepackage{listings}
\usepackage{mhchem}
\def\BibTeX{{\rm B\kern-.05em{\sc i\kern-.025em b}\kern-.08em
    T\kern-.1667em\lower.7ex\hbox{E}\kern-.125emX}}

\usepackage{tikz}

\newcommand\copyrighttext{%
  \footnotesize \textcopyright \the\year{} IEEE. Personal use of this material is permitted. Permission from IEEE must be obtained for all other uses, in any current or future media, including reprinting/republishing this material for advertising or promotional purposes, creating new collective works, for resale or redistribution to servers or lists, or reuse of any copyrighted component of this work in other works.}

\newcommand\copyrightnotice{%
\begin{tikzpicture}[remember picture,overlay]
\node[anchor=south,yshift=10pt] at (current page.south) {\fbox{\parbox{\dimexpr0.75\textwidth-\fboxsep-\fboxrule\relax}{\copyrighttext}}};
\end{tikzpicture}%
}
    
\begin{document}

\title{
Visual Cooperative Drone Tracking for Open-Path Gas Measurements\\
\thanks{This work was funded by project SPP2433 from the German Research Foundation.}
}

\author{\IEEEauthorblockN{Marius Schaab\IEEEauthorrefmark{1}, Alisha Kiefer\IEEEauthorrefmark{1}, Thomas Wiedemann\IEEEauthorrefmark{1}\IEEEauthorrefmark{2}, Patrick Hinsen\IEEEauthorrefmark{2} and Achim J. Lilienthal\IEEEauthorrefmark{1}}
\IEEEauthorblockA{\IEEEauthorrefmark{1}
\textit{Technical University of Munich (TUM)}, Munich, Germany \\
\IEEEauthorrefmark{2}\textit{German Aerospace Center (DLR)}, Weßling, Germany \\
Correspondence: marius.schaab@tum.de}
}

\maketitle

\copyrightnotice

\begin{abstract}
Open-path Tunable Diode Laser Absorption Spectroscopy offers an effective method for measuring, mapping, and monitoring gas concentrations, such as leaking \ce{CO2} or methane.
Compared to spatial sampling of gas distributions using in-situ sensors, open-path sensors in combination with gas tomography algorithms can cover large outdoor environments faster in a non-invasive way.
However, the requirement of a dedicated reflection surface for the open-path laser makes automating the spatial sampling process challenging.  
This publication presents a robotic system for collecting open-path measurements, making use of a sensor mounted on a ground-based pan-tilt unit and a small drone carrying a reflector.
By means of a zoom camera, the ground unit visually tracks red LED markers mounted on the drone and aligns the sensor's laser beam with the reflector.
Incorporating GNSS position information provided by the drone's flight controller further improves the tracking approach.
Outdoor experiments validated the system's performance, demonstrating successful autonomous tracking and valid \ce{CO2} measurements at distances up to 60 meters. 
Furthermore, the system successfully measured a \ce{CO2} plume without interference from the drone's propulsion system, demonstrating its superiority compared to flying in-situ sensors. 
\end{abstract}


\section{Introduction}
The leakage of gases in industry or from natural disasters, like volcanic eruptions or  forest fires,  pollutes our air, posing a risk to human health and the global climate. 
The latest global greenhouse gas emission report from the European Commission revealed a new record representing the highest level of \ce{CO2}$_{\mathrm{eq}}$ emission recorded\cite{EU_GHG}.
To determine the extent and locations of emissions, it is essential to measure and observe the airborne gas concentration. 
Precise measurements help to infer leakages and develop mitigation strategies. 
However, leaking gas disperses in a chaotic pattern, creating large local concentration differences in space. 
Measuring these concentrations with in-situ sensors shows several disadvantages:
(i) Dense spatial sampling is time-consuming. (ii) In-situ sampling carries the risk that high-concentration spots may be missed. (iii) Placing in-situ sensors requires contact with the gas plume and might disturb the gas dispersion mechanism, causing unreliable measurements -- especially when dispatching a drone.

\begin{figure}
    \centering
    \includegraphics[width=0.9\linewidth]{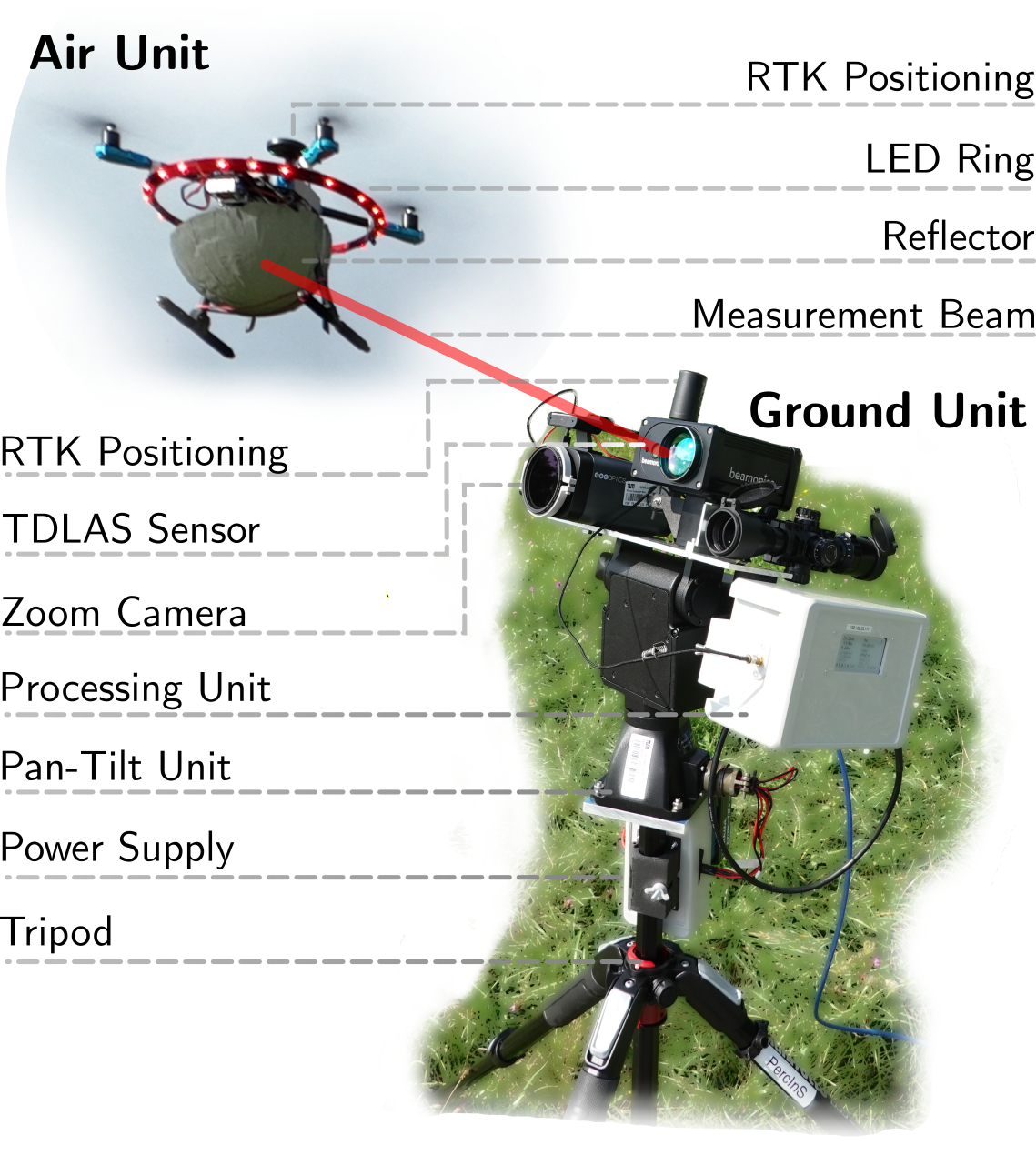}
    \caption{Our measurement system consists of a drone carrying a reflector and a ground station with the open-path sensor.}
    \label{fig:measurement_sys}
\end{figure}

Gas tomography \cite{drescher, thomas, marius} solves these issues by fusing multiple open-path measurements to reconstruct the gas distribution. 
An  open-path sensor, such as a Tunable Diode Laser Absorption Spectroscopy (TDLAS) system, measures gas concentration 
by sending a laser beam of a specific wavelength through an analyte gas.
By analyzing the reflected light with respect to absorption, the sensor determines the gas concentration in parts per million times meter (\unit{ppm*\metre}), which corresponds to the integral of the gas concentration along the laser path.
However, this principle requires a dedicated reflector.
At short distances, up to ten meters, the reflector can be something simple, such as cardboard or white-painted walls, or similar features already available in the environment.
For distances exceeding ten meters, the TDLAS requires a specialized reflector that must be mobile to accommodate multiple measurement constellations.
Choosing the position of the sensor and reflector freely has the benefit that neither needs to be in physical contact with the gas plume, such that measurements do not affect the gas dispersion. 
This work addresses the challenges in acquiring automated open-path gas measurements for gas tomography.  
We investigate the design, performance of a robotic measurement system and validate it by measuring a \ce{CO2} plume released from a gas cylinder in a field experiment.

The idea of utilizing robotic systems for open-path measurements is not new.
Bennetts et al. (2014) presented a ground-based robot for gas tomography measurements \cite{Bennetts}. 
Later, Neumann et al. (2018) extended this approach to drones carrying the TDLAS \cite{Neumann}. 
Both methods used natural reflectors in the environment. 
The natural reflectors impose constraints on the sensor's positioning. The sensor must be positioned in close proximity to one of the reflectors. 
Lohrke et al. (2025) introduced a system based on two ground-based robots \cite{Lohrke},
one carrying the sensor and one carrying the reflector. This method leads to an alignment problem, such that the TDLAS beam hits the reflector.
Lohrke addressed the problem by introducing a camera-based tracking system and demonstrated successful alignment in an outdoor experiment in an area of \qtyproduct{15 x 7}{\metre}.
The two robots measure gas concentration with freely selectable positions for the sensor and reflector near the ground.
Measuring a gas plume in the air a few meters above the ground is not possible.

To this end, we propose mounting the reflector on a drone tracked by a Pan-Tilt Unit (PTU) that carries the TDLAS sensor and automatically aligns the laser with the reflector.
The cooperative tracking system operates using a zoom camera to visually follow red LED markers attached to a ring on the drone. 
Furthermore, the drone communicates the GNSS position to the PTU, making the tracking more robust under adverse visual conditions. Fig.~\ref{fig:measurement_sys} shows our measurement system.
The system is capable of automated tracking and measurements up to \qty{60}{\metre} in a three-dimensional space. 

To summarize our core contribution, we propose
\begin{itemize}
    \item a robust design of a data acquisition system for automating open-path measurements,
    \item cooperative drone tracking based on camera and GNSS data, surpassing the \qty{60}{\metre} sensor range,
    \item validation in outdoor experiments using a low-emission \ce{CO2} source.
\end{itemize}

\section{System Design}

Our system, shown in Fig.~\ref{fig:measurement_sys}, consists of two main units: an air unit -- a drone equipped with a reflector and LED markers -- and a ground unit based on a PTU that carries the TDLAS sensor and a zoom camera.
A RTK-GNSS receiver provides precise localization for both units, and WiFi enables communication between the units. 
The drone is a Holybro X500 with a Pixhawk Cube Orange flight controller. 
The open-source flight controller can fly preplaned routes, as well as adapt to new position commands mid-flight. 
The drone carries an on-board Raspberry Pi 4 that receives RTK-GNSS correction signals via WiFi and forwards them to the flight controller.
In addition, the Raspberry Pi 4 can also send and request flight parameters and forward them to the ground unit.
The drone carries a red LED ring for easy visual detection and a hemispherical retroreflector beneath. 
The reflector is made of styrofoam, has a diameter of \qty{25}{\centi\metre}, and is covered with retroreflective fabric.
The PTU, mounted on a tripod, carries the TDLAS sensor and a zoom camera. 
The PTU, type FLIR PTU-D38E, has a pan/tilt resolution of \ang{0.006}/\ang{0.003} and supports speed or position control commands over a serial interface, enabling precise laser alignment of \qty{1}{cm} at \qty{100}{\metre} distance.
The camera is a PTZOptics Studio Pro, featuring a 12x optical zoom and a USB interface for video and control transfer. Using the zoom feature, we can spot the drone even if it’s flying at a distance of \qty{100}{\metre}.  We operate the camera at a resolution of \qtyproduct{640 x 480}{px} for faster image processing. 
As an open-path gas sensor, we are using a BeamSight TDLAS system from the company Beamonics. 
The TDLAS system measures \ce{CO2} based on absorption using a narrow-band infrared laser at a wavelength of \qty{2.05}{\micro\metre}. 
The laser shoots a beam through an analyte gas, which is then reflected, and the returning light intensity is measured at the sensor.
By varying the operating current of the laser diode, the laser can shift its frequency by a few nanometers. The sensor takes multiple measurements at different frequencies and generates a narrow-banded spectrogram. 
The sensor's onboard processor translates this spectrogram into an integral of the \ce{CO2} concentration along the laser beam.
The measured variable is expressed in \unit{ppm*\metre}. By knowing the position of the sensor and the reflector, we can calculate the average \ce{CO2} concentration in \unit{ppm} along the beam. 
In addition to the concentration, the sensor returns a measure of the signal strength and quality, as well as an internal status code. 
The status code reflects whether a measurement was successful (OK) or an error occurred, e.g., when the system detects no returning signal, either because the sensor is misaligned with the reflector or because the distance to the reflector is too high.
The sensor also returns an error when high external light pollution occurs, such as when it points directly at the sun.
Further, the sensor reports warnings when a measurement is still possible but the returning light intensity gets weak, or a high transmission warning when strong reflections start to overexpose the receiver. 
Our TDLAS sensor measures exclusively \ce{CO2}, but we can adapt the measurement system to other gases, such as methane or \ce{H2S}. We would need a new sensor with a laser diode and detector specifically designed to fit the absorption band of the gas of interest.

\begin{figure*}[htp]
    \centering
    \includegraphics[width=1.0\linewidth]{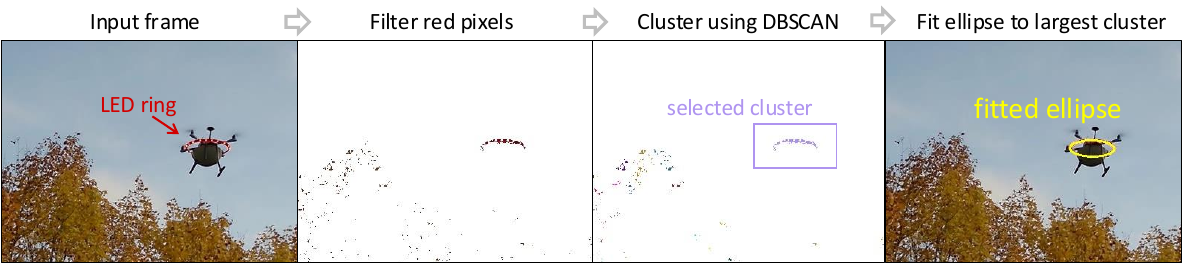}
    \caption{Processing steps to determine the drone position in the frame.}
    \label{fig:frames}
\end{figure*}

\section{Methodology}
This section covers two steps. The first one describes our drone tracking approach, which aligns the TDLAS laser beam with the reflector on the drone, and the second one processes the measured values to determine the concentration of \ce{CO2}. 

\subsection{Tracking Method}
The drone tracking primarily works based on visual detection of the drone in a camera frame.
To support reliable detection over a wide range of distances, an automatic camera zoom control dynamically adjusts the field of view, ensuring that the drone remains visible and properly scaled in the image.
The drone also communicates its location to the tracking unit to catch errors caused by vision loss. 
A PI controller converts a missalignment into velocity commands for the PTU. All computation runs on a Raspberry Pi 5 attached to the camera and PTU.

Fig.~\ref{fig:frames} depicts the individual processing steps.
We localize the drone in the camera image on a single frame basis. 
We convert the RGB color space of the image into an HSV color space. We filter for red pixels within the HSV color space.
The advantage of the HSV (Hue, Saturation, Value) color space is that it approximately decouples the color information (Hue and Saturation) from the brightness (Value). This enhances the robustness and stability of our tracking algorithm against varying lighting conditions, including shadows and highlights. 
To compensate for background noise, we introduce the Density-Based Spatial Clustering of Applications with Noise (DBSCAN) algorithm \cite{DBSCAN}. This density-based clustering algorithm groups closely packed red pixels (dense regions) together, marking red pixels in low-density regions as outliers (noise).
By selecting the largest group, we get the pixels that belong to the LED ring of the drone. Note that we assume here that our drone is the largest red object in the frame. 
To determine the pixel position $\bm{p}^\text{drone}$ of the drone, we use OpenCV's \lstinline|fitEllipse| function. This function implements the first algorithm described in \cite{ellipse}. 
 The center of the ellipse corresponds to the position $\bm{w}$ of the drone relative to the center pixel $\bm{p}_\text{center}$ and normalized with the image resolution:
\begin{equation}
    \bm{w} = \begin{pmatrix}
        \left(p^\text{drone}_{x_1} - p_{x_1}^{\text{center}}\right) / \text{res}_{x_1} \\
        \left(p^\text{drone}_{x_2} - p_{x_2}^{\text{center}}\right) / \text{res}_{x_2}
    \end{pmatrix}.  
\end{equation}

To support long-distance tracking and measurement, our method includes adaptation of the camera zoom according to the size of the ellipse. If the size of the drone falls below a lower limit, we optically zoom in, and if the drone exceeds an upper limit, we zoom out. 
If the drone leaves the camera frame, such that there are no more red pixels in the frame, we also zoom out. 
It is important to note that by adjusting the zoom, we also change the horizontal and vertical field-of-view angles, $\alpha_\text{HFOV} $ and $\alpha_\text{VFOV}$. 

The angles $\alpha_\text{HFOV} $ and $\alpha_\text{VFOV}$ are needed to calculate the horizontal offset angle between the camera center, i.e., the current PTU orientation, and the direction to the drone:
\begin{equation}
    \Delta\phi = \arctan\left( 2 \cdot w_1 \cdot \tan(0.5 \cdot \alpha_\text{HFOV})\right),
\end{equation}
and analog to that, the vertical offset angle
\begin{equation}
    \Delta\theta = \arctan\left( 2 \cdot w_2 \cdot \tan(0.5 \cdot \alpha_\text{VFOV})\right).
\end{equation}

$\Delta\phi$ and $\Delta\theta$ describe the current misalignment in the drone tracking. 
We employ a PI controller to control the PTU.
The controller minimizes misalignment by sending velocity commands to the PTU, thereby aligning the TDLAS laser beam with the reflector on the drone. 

In addition, we make use of GNSS position information from the drone to check if the drone leaves the camera's field of view.
Rapid changes of the drone's flight direction or flight routes exceeding the tilt limits of the PTU could lead to vision loss. 
In such a case, we calculate the missalignment angles from the GNSS position of the drone until the cooperative controller brings the drone back into view. 
This procedure makes our tracking fail-safe against potential vision loss. 

\subsection{Gas Measurements}
Our tracking method ensures the alignment between the TDLAS sensor and the reflector while the drone flies a predefined route during an experiment. 
To make use of the TDLAS data in postprocessing, we require two data streams: 
First, the drone logs its current position $\bm{x}^\text{drone}_j$ and time $t_j$. 
Second, the ground unit records measurements $m_i$, TDLAS status codes, and GNSS position of the TDLAS $\bm{x}^\text{TDLAS}_i$ as well as time $t_i$.
The measurements of the TDLAS are the integral over the real gas concentration $u^*(\bm{x}, t_i)$:
\begin{equation}
    m_i = \int_0^{1} u^*
    \!
    \left(
    (\bm{x}^\text{drone}_i - \bm{x}^\text{TDLAS}_i) s+ \bm{x}^\text{TDLAS}_i, \, t_i
    \right) 
    \, \mathrm{d}s
    \, .
\end{equation}
Note that we linearly interpolate the drone's recorded position at time steps $t_j$ to obtain its position $\bm{x}^\text{drone}_i$ at time step $t_i$.
To calculate the average \ce{CO2} concentration of each measurement along the laser beam, we compute
\begin{equation}
    \bar{u}_i = \frac{m_i}{d_i},
\end{equation}
where $d_i$ is the Euclidean distance between $\bm{x}^\text{TDLAS}_i$ and $\bm{x}^\text{drone}_i$.
To this end, we convert the position recorded in the Geographic Coordinate System (GCS) with latitude and longitude coordinates into a Cartesian coordinate system using the appropriate zone of the Universal Transverse Mercator (UTM) system. 

Our procedure for calculating gas concentrations includes a few simplifications:
The recorded TDLAS position refers to the position of the GNSS antenna, not the position of the laser transmitter-receiver. 
They are spaced approximately \qty{10}{\centi\metre} apart from each other. 
The recorded drone position is the position of the GNSS antenna fused with data from an on-board inertial measurement unit. 
In the worst case, the antenna and reflection point are \qty{40}{\centi\metre} apart from each other. 
We also neglect the positioning accuracy of the RTK-GNSS receiver, which is according to the datasheet within \qty{\pm 1}{\centi\metre}\cite{ublox}. 
For a background concentration of \qty{400}{ppm} and a measurement distance of \qty{50}{\metre}, these simplifications introduce a worst-case error of \qty{4}{ppm}. 
According to the manufacturer, our TDLAS sensors' accuracy is below \qty{1}{ppm} at \qty{50}{\metre} (\qty{40}{ppm*\metre}). Addressing these offsets would improve the performance of our system in the future.

\section{Experiments}
We conducted two field experiments. The first experiment evaluates the tracking performance in terms of distance. The second experiment encompasses measurements obtained from a low-emission gas source.
In both experiments, we set the flight speed to \qty{1}{\metre\per\second}. 
For evaluating the tracking performance, the drone flew a pre-programmed zig-zag route, depicted in Fig.~\ref{fig:path}. 
The drone did not stop at a waypoint and continuously increased its altitude from \qty{0.8}{\metre} to \qty{7.9}{\metre} above the TDLAS position.
To evaluate the performance of the tracking, we have a look at the status codes of the TDLAS sensor that indicate a successful measurement.

In the second experiment, we placed a gas source \qty{16}{\metre} south of the TDLAS. The gas source was a \ce{CO2} bottle with a release tube attached approximately \qty{1.5}{\metre} above ground.
A pressure and flow regulator was used to release \ce{CO2} at a rate of \qty[per-mode = symbol]{25}{\litre\per\minute}. 
For the drone, we planned a flight route around \qty{30}{\metre} further south of the gas source on lines iterating from left to right and back relative to the source, while slowly lowering its altitude from \qty{6}{\metre} down to \qty{1.8}{\metre}.
Using our proposed drone tracking system, the TDLAS scans a whole z-x plane, creating an image of the invisible \ce{CO2} plume. 
It was cloudy during both experiments; thus, the interference of sunlight on the TDLAS could be ignored.

\section{Results}

\begin{figure}
    \centering
    \includegraphics[width=0.99\linewidth]{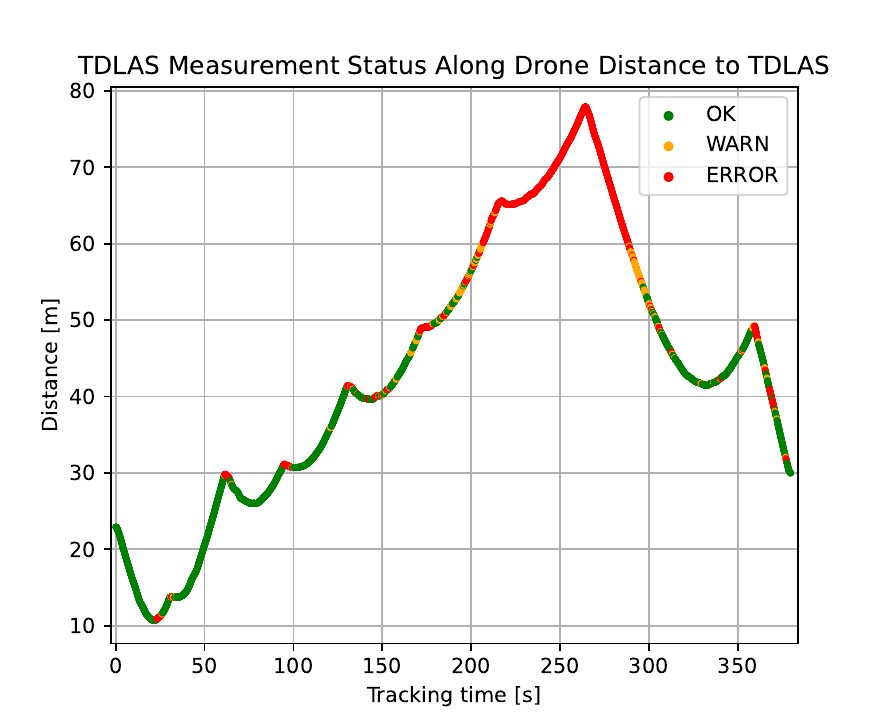}
    \caption{Evaluation of the distance measurements consistency. The TDLAS achieved valid measurements up to a distance of \qty{60}{\metre}.}
    \label{fig:distance}
\end{figure}

The tracking experiment shows successful alignment between the TDLAS and the reflector. 
The status code provided by the TDLAS serves as a performance indicator for successful measurements. Status code OK and WARNing are considered as valid measurements, while status code ERROR indicates an invalid measurement, either due to misalignment of the TDLAS sensor to the reflector or due to a too large distance between sensor and reflector.
Fig.~\ref{fig:distance} plots the distances between TDLAS and the drone and status codes over time for the zig-zag route. 
We see that the system achieves valid measurements up to \qty{60}{\metre}. 
Above that, we do not obtain any valid measurements, likely due to the small size of the reflector, which reflects too little light back to the sensor.
Previous experiments, using manual alignments, achieved similar performance.
Increasing the reflector size would increase the distance performance, but would decrease flight time due to a heavier payload. 
\begin{figure}
    \centering
    \includegraphics[width=0.99\linewidth]{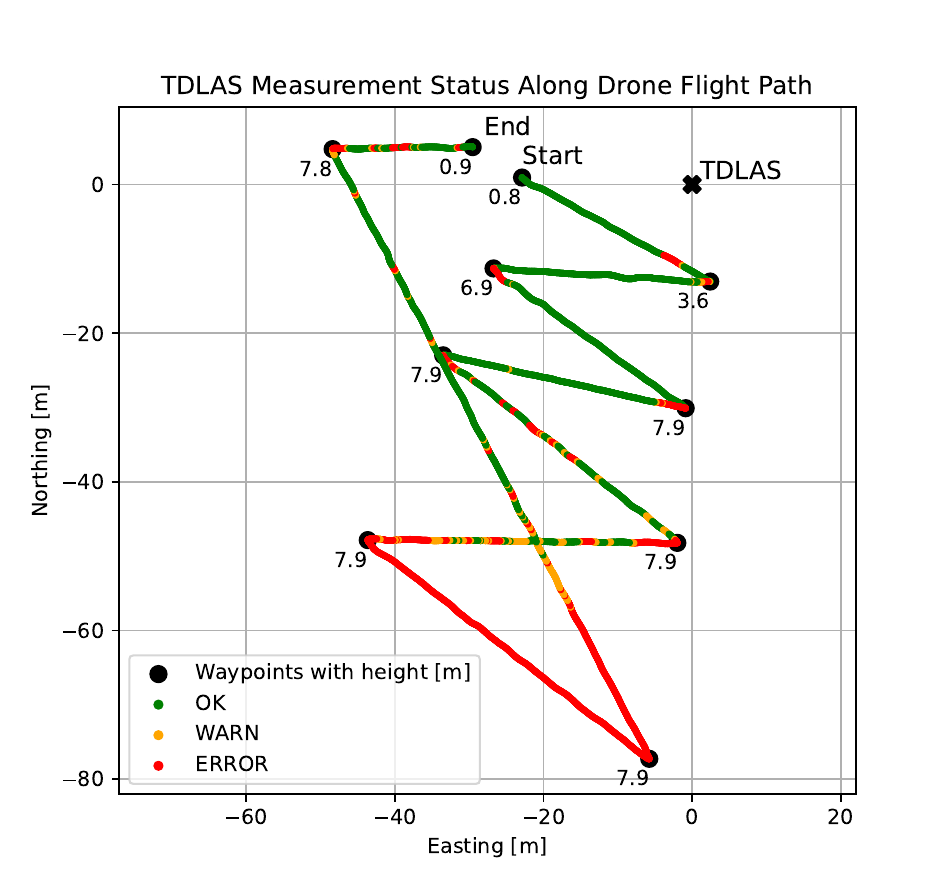}
    \caption{Benchmark flight route for evaluating the performance of the tracking algorithm. Measurements with status codes OK and WARN are valid. TDLAS is the position of our sensor.}
    \label{fig:path}
\end{figure}
Fig.~\ref{fig:path} shows the measurements on the previously mentioned zig-zag route. 
The PI controller needs a short time to adapt to abrupt changes in flight direction.
Therefore, we shortly lost track when the drone changed direction at the waypoints.  
Another point where the PI controller lost track is on the first diagonal path after the start. 
As the drone passes close by the TDLAS, the speed of the PTU is not fast enough to follow the drone.
Reducing the flight speed would improve the tracking performance, but also increase measurement time. Also, a preplanned flight path that takes the capabilities of the PTU into account could improve tracking performance, e.g. by using smooth transitions between different flight directions.

\begin{figure}
    \centering
    \includegraphics[width=0.95\linewidth]{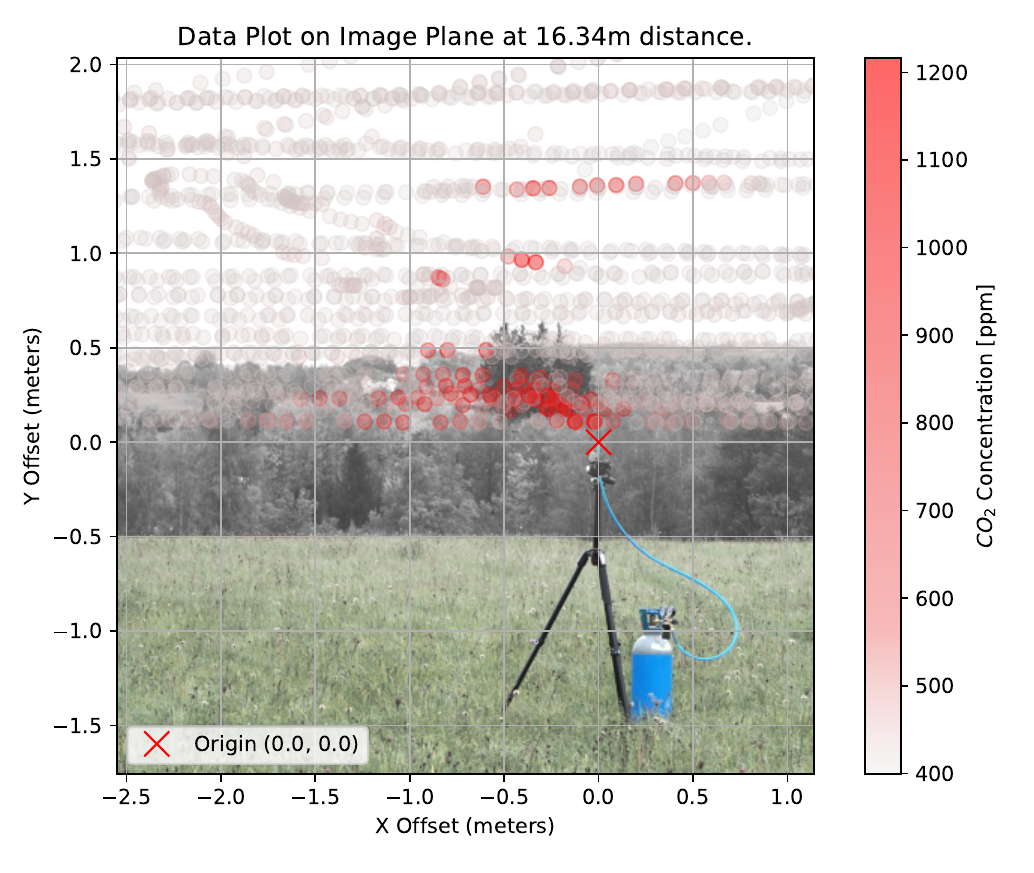}
    \caption{Calculated gas concentrations from three flights using a \ce{CO2} bottle with a release rate of \qty[per-mode = symbol]{25}{\litre\per\minute}.}
    \label{fig:gas_image}
\end{figure}

Fig.~\ref{fig:gas_image} plots measurements of the second experiment as described in the previous section onto a south-facing picture of the gas source.
Each measurement is projected as a point onto the image plane passing through the source. The color describes the average concentration along the beam. 
We observed the main \ce{CO2} plume travelling south-east (into the image and to the left), which matches the prevailing wind direction. 
Vertically, the plume's behaviour was dominated by its physical properties. 
As the \ce{CO2} was released from the pressurised bottle, it cooled significantly; this cooling, combined with \ce{CO2} being heavier than air, caused the plume to sink rather than rise.
Despite this tendency to sink, the TDLAS also measured \ce{CO2} concentrations \qty{1.4}{\metre} above the gas source. The measurements in Fig.~\ref{fig:gas_image} were distributed over a 36-minute period, during which the wind was not consistent. The wind changed the intensity and direction.
Therefore, these higher-altitude readings may not be due to wind; they could be attributed to the initial inertia from the gas release pushing the plume upward during temporary wind lulls. 
The results show that our system is sensitive enough to capture even these low and complex gas concentrations. 
The downwash from our drone-based measurement systems does not affect the shape of the plume, since the drone is placed at a distance behind the source.

\section{Conclusion}
We developed and validated an automated robotic system for non-interceptive, open-path \ce{CO2} measurements. 
The system employs a ground-based PTU to automatically align a TDLAS laser beam with a reflector-equipped drone.
To this end, we make use of LED markers mounted on the drone that are visually tracked by the camera.
This approach was successfully demonstrated to be able to remotely measure a \ce{CO2} plume without the sensor or drone interfering with the plume dispersion.
The system operates autonomously and achieves valid measurements at distances up to \qty{60}{\metre}.

While a larger reflector would, on the one hand, extend the range of our measurement, it would, on the other hand, reduce the drone's flight time due to higher power consumption for lifting the heavier payload. The reflector with a \qty{25}{cm} diameter serves therefore as a good tradeoff between measurement time and range. 
Additionally, the system's overall precision is currently limited by physical offsets between components, although RTK-GNSS provides high accuracy. Offsets of up to \qty{40}{cm} between the drone's GNSS antenna and the reflection point introduce a worst-case error of approximately \qty{4}{ppm} at \qty{50}{\metre}. This error exceeds the TDLAS sensor’s inherent accuracy of less than \qty{1}{ppm}. Compensating these offsets requires keeping track of the precise orientation of the PTU and the drone, such that we can calculate the reflection point at the reflector surface. Including the orientation would add complexity to our method, but it would improve the accuracy. 
Also, measuring the gas concentration of a dynamic plume is quite challenging. Changing wind conditions and the chaotic dispersion of gas lead to high spatial and temporal heterogeneity. A single measurement captures only a snapshot of the gas concentration. Common applications, such as leak localization or emission estimations, require the average concentration along a measurement path. We would need to sample for a long time at a single measurement location to reduce the uncertainty, or use more advanced measurement techniques, such as gas tomography.

In the future, we will utilize the system to reconstruct the full plume shape using gas tomography. Our drone-based measurement system enables us to capture data for a three-dimensional reconstruction of a plume. 
Furthermore, we will extend the system's capabilities by mounting the PTU on a moving robotic platform, allowing more flexible changes of sensor-reflector constellations.

\section*{Acknowledgment}

The authors thank the German Research Foundation (DFG) for funding and support of this research through the SPP2433 project. The authors also thank the Munich Institute for Robotics and Machine Intelligence (MIRMI) at the Technical University of Munich for their support.

\end{document}